\documentclass{article}
\newcommand{\final}{0}

\usepackage{times,color}
\usepackage[table]{xcolor}
\usepackage{parskip}
\usepackage[pdftex]{graphicx}
\usepackage{ifthen}
\usepackage[hidelinks]{hyperref}
\usepackage{float}
\usepackage{morefloats}
\usepackage{alltt}
\usepackage{amsmath}
\usepackage{amssymb}
\usepackage{amsthm}
\usepackage{ulem}
\usepackage{mathrsfs}
\usepackage{newlfont} 
\usepackage{ulem}
\usepackage{floatflt}
\usepackage{wrapfig}
\usepackage{fixltx2e}
\usepackage{subfig} 
\usepackage{multirow}
\usepackage{cleveref}
\Crefname{equation}{Eq.}{Eqs.}
\Crefname{figure}{Fig.}{Figs.}
\usepackage[T1]{fontenc} 
\usepackage{etoolbox}
\usepackage{xfrac}
\usepackage[export]{adjustbox}
\usepackage{array}
\usepackage{tikz}
\usetikzlibrary{calc,tikzmark}
\usepackage{spconf} 
\usepackage[percent]{overpic}

\newcommand{\eg}{{e.g.~}}

\definecolor{YodaColor}{rgb}{0,0.3,0} 
\definecolor{ChiakaiColor}{rgb}{0,0,0.8} 
\definecolor{JanaColor}{rgb}{0,0.5,0.8} 
\definecolor{FuhaoColor}{rgb}{0.8,0,0.8} 
\newcommand{\ckliang}[1]{{\color{ChiakaiColor} Chia-Kai: #1 $\qed$}}
\newcommand{\fhshi}[1]{{\color{FuhaoColor} Fuhao: #1 $\qed$}}
\newcommand{\reviewer}[2][]{{\color{YodaColor} Reviewer#1: #2 $\qed$}} 

\newcommand{\warning}[1]{{\it\color{red} #1}}
\newcommand{\note}[1]{{\it\color{blue} #1}}
\newcommand{\nothing}[1]{}

\definecolor{AudioColor}{rgb}{0.56,0.34,0.62}
\newcommand{\audio}[2][]{{\color{AudioColor} Audio: #2 $\qed$}}
\definecolor{VideoColor}{rgb}{0.44,0.66,0.38}
\newcommand{\video}[1]{{\color{VideoColor} Video: #1 $\qed$}}
\definecolor{DeadlineColor}{rgb}{0.9,0.4,0}
\newcommand{\deadline}[1]{{\bf\color{DeadlineColor} ETA: #1}}

\definecolor{OldColor}{rgb}{0.5,0.5,0.5}
\newcommand{\old}[1]{{\color{OldColor} #1}}
\definecolor{NewColor}{rgb}{0.9,0.4,0}

\definecolor{DeleteColor}{rgb}{0.1,0.6,1.0}
\newcommand{\delete}[1]{{\color{DeleteColor} #1}}
\definecolor{MoveColor}{rgb}{0.5,0.1,0.5}

\definecolor{figred}{rgb}{1,0,0}
\definecolor{figgreen}{rgb}{0,0.6,0}
\definecolor{figblue}{rgb}{0,0,1}
\definecolor{figpink}{rgb}{1,0.63,0.63}

\ifthenelse{\equal{\final}{1}}
{
\renewcommand{\ckliang}[1]{}
\renewcommand{\fhshi}[1]{}
\renewcommand{\reviewer}[2][]{}

\renewcommand{\warning}[1]{}
\renewcommand{\note}[1]{}
\renewcommand{\old}[1]{}
\renewcommand{\audio}[2][]{}
\renewcommand{\video}[1]{}
\renewcommand{\deadline}[1]{}
}
{}

\newcommand{\pseudocode}{Pseudocode}
\floatstyle{plain}
\newfloat{algorithm}{tbhp}{lop}
\floatname{algorithm}{\pseudocode}

\ifthenelse{\isundefined{\diff}}
{

\renewcommand{\delete}[1]{}

}
{
\ifthenelse{\equal{\diff}{0}}
{

\renewcommand{\delete}[1]{}

}
{}
}

\ifdefined\google

\else

\fi

\newcommand{\filename}[1]{\url{#1}}
\newcommand{\foldername}[1]{\url{#1}}
\newcommand{\email}[1]{\url{#1}}

\catcode`_=12
\begingroup\lccode`~=`_\lowercase{\endgroup\let~\sb}
\mathcode`_="8000

\newcommand{\colvec}[1]{\mathbf{#1}}
\newcommand{\mat}[1]{\mathbf{#1}}
\def\landmark{\colvec{L}}
\def\headcenter{\colvec{H}}
\def\landmarkmean{\colvec{C}}

\def\virtual{v}
\def\real{r}
\def\pose{\mathcal{P}}
\def\rotation{\mat{r}}
\def\Rotation{\mat{R}}
\def\trans{\colvec{t}}
\def\intrinsic{\mat{K}}
\def\focal{f}
\def\homography{\mat{\Pi}}

\def\prev{\! -1}
\def\prevprev{\! -2}

\def\solidangle{\Omega}
\def\argmin{\mathrm{argmin}}
\def\logistic{\mathtt{logistic}}
\def\protrude{\mathtt{protrude}}
\def\proj{\mathtt{proj}}
\newcommand{\ltwo}[1]{\|#1\|_2}

\def\fit{f}
\def\distortion{d}
\def\follow{o}
\def\smoothr{r}
\def\smootht{t}
\def\protrusion{p}

\graphicspath{
    {figs/}
}

\title{Steadiface: Real-Time Face-Centric Stabilization on Mobile Phones}
\name{Fuhao Shi, Sung-Fang Tsai, Youyou Wang, and Chia-Kai Liang}
\address{Google Inc.}

\begin{document}
\normalem
\ninept 
\maketitle

\begin{abstract}
\vspace{0.2in}
We present \emph{Steadiface}, a new real-time face-centric video stabilization method that simultaneously removes hand shake and keeps subject's head stable.
We use a CNN to estimate the face landmarks and use them to optimize a stabilized head center.
We then formulate an optimization problem to find a virtual camera pose that locates the face to the stabilized head center while retains smooth rotation and translation transitions across frames.
We test the proposed method on fieldtest videos and show it stabilizes both the head motion and background.
It is robust to large head pose, occlusion, facial appearance variations, and different kinds of camera motions.
We show our method advances the state of art in selfie video stabilization by comparing against alternative methods.
The whole process runs very efficiently on a modern mobile phone (8.1 ms/frame).
\end{abstract}

\section{Introduction}
\label{sec:intro}

Stable appearance of human faces is crucial for selfie videos, live shows, or vlogging, which all are now popular features on mobile phones.
However, unintentional hand shakes and head translations during recording can easily make the face unstable.
Unfortunately, most existing video stabilization methods do not stabilize selfie videos well.
The face motion, which is usually very different from the background, can reamian unstable after stabilization.

We present \emph{Steadiface}, a real-time gyro-based face-centric video stabilization method that simultaneously removes hand shake and keeps head stable. Moreover, it runs real-time on the mobile phone without delaying the video stream, and provides the WYSIWYG user experience.
The proposed method uses the gyroscope to obtain the camera motions and an efficient CNN to extract the face landmarks.
We then formulate an optimization problem to jointly stabilize the camera and head motion.
We also dynamically control the weighting in the optimization process for robustness.

\emph{Steadiface} is highly efficient: it takes only 8.1 ms/frame on Google Pixel 3 (Qualcomm Snapdragon 845 CPU, Adreno 630 GPU).
We tested it on many videos with challenging head poses, occlusions, and camera motions, and obtained good results in all cases \cite{Shi:2019:SUP}.

This work has made the following contributions:
\vspace{-0.07in}
\begin{itemize}
	\item The first real-time end-to-end stabilization system that simultaneously stabilizes both head and camera motion.
	\item A novel algorithm that combines both face and gyro stableness into a single objective function for joint optimization.
	\item An effective weight adjustment scheme robust to head pose variance and noisy landmark locations.
    \item We perform extensive comparisons between our method and the state-of-the-art ones.
\end{itemize}

\begin{figure}[t]
    \centering
    \includegraphics[width=0.9\columnwidth]{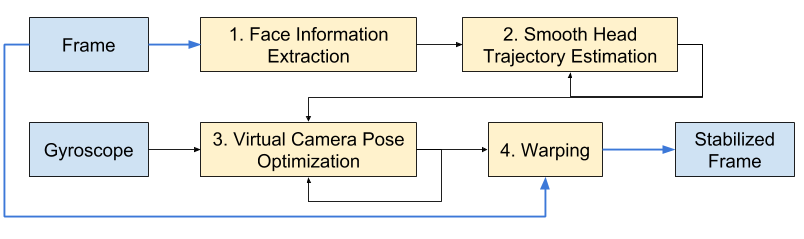}
    \vspace{-0.15in}
    \caption{\emph{Steadiface} algorithm pipeline.
        \label{fig:overview}
    }
\end{figure}

\section{Background}
\label{sec:background}
The electronic video stabilization systems usually consist of three components: motion estimation, motion compensation and image composition \cite{Morimoto:1998:EIS}.
There are two popular  motion estimation methods: image based and sensor based.
Given the estimated motion profiles, motion compensation creates a smooth motion via filtering \cite{Karpenko:2011:DVS} or optimization \cite{Grundmann:2011:ADV}, and image composition adjusts the input video into a stabilized one via shifting or warping.
Modern methods also handle rolling shutter distortions during warping \cite{Liang:2008:ACR,Grundmann:2011:ADV,Karpenko:2011:DVS}.

There are also non-electronic video stabilization methods, such as mechanical gimbal or the optical image stabilizer (OIS).
However, they do no work for face centric videos and we skip them here for brevity.

\textbf{Sensor-based stabilization} methods use gyroscope, OIS or their combinations to model the camera model as rotation and translation, and stabilize the videos by smoothing the virtual pose changes \cite{Karpenko:2011:DVS,Bell:2014:ANF,Liang:2017:FVS}.
Our work is mostly related to the fused video stabilization as used in Google Pixel 3 \cite{Liang:2017:FVS}.
Similarly, we extract the gyroscope signal to integrate as the camera pose, and warp the frame by dividing the input frame into a mesh and warp each part separately to handle the rolling shutter distortion \cite{Liang:2008:ACR}.
Our key novelty is to detect faces and tracks the facial landmarks from the input frame, and fuse both gyro and face information to estimate the best joint face and background stability.

\textbf{Image-based stabilization} methods detect/track the features across video frames, and stabilize the motion by smoothing the camera path \cite{Liu:2009:CPW,Goldstein:2012:VSE,Liu:2011:SVS,Liu:2013:BCP}.
As the feature tracking is noisy or camera motion estimation can be difficult in degenerated cases, most methods focus on improving the robustness.
However, they are not designed to handle dominant moving subjects like faces.

Yu and Ramamoorthi proposed an image-based method for face-centric stabilization \cite{Yu:2018:SVS}.
The head motion is modeled as the 3D head center of the reconstructed head, and the background motion is tracked by dense optical flow.
An optimization for homography and additional mesh adjustment is performed to obtain the best trade-off between face and background stability.
However, their approach is slow and sensitive to motion estimation errors.
The required 3D face reconstruction, despite the advances from 3D fitting using 2D landmarks\,\cite{Blanz:1999:Morphable, Shi:2014:Automatic, Thies:2016:face2face, Garrido:2016:3DFaceRig} to direct regression/CNN inference\,\cite{Cao:2014:DDE, Jourabloo:2016:CNN, Tewari:2017:Mofa}, can be either slow or power-consuming for the video recording task on a mobile phone.
\nothing{\ckliang{We may ping them before showing their results}}
\nothing{
\ckliang{Cite some papers to give an example on how difficult the head geometry estimation is?}
\fhshi{Added. May refine the wording.}}

Our method is different from theirs in four parts.
First, we use 2D facial features to represent the head motion without expensive 3D head reconstruction.
Second, we use gyroscope to obtain the camera motion and do not require feature tracking or optical flow.
Third, we use a two-step process to explicitly control the head stability and avoid error accumulations.
We first estimate a stabilized head trajectory and then optimize the virtual camera pose to align the face to that trajectory.
Finally, we use a novel metric for head and background stability with dynamic weight adjustment to handle pose variations and noisy landmarks.

\section{Real-time video stabilization}
\label{sec:realtime}
\emph{Steadiface} takes the input video frame as well as gyroscope readout as inputs, and outputs a warping mesh that warps the original video frame to the stabilized result.
The overall pipeline is shown in \Cref{fig:overview}.
First, face information is extracted from the input video frame, including face bounding box and 2D facial landmarks.
We then estimate a smooth target head trajectory from the landmark locations.
A joint optimization then takes the gyroscope, face information, and head trajectory to find the optimal vitrual camera pose.
Finally the warping is aplied to transform the input frame to the stabilized one.

All processings are performed at each frame sequentially, and in the following discussions, we will drop the frame index or time value for brevity. We will describe the first two steps in this section.

\subsection{Face Information Extraction}
This step takes the video frame as input, detects the face bounding boxes, select the best face to process, and then infers its facial landmarks.

\setlength{\columnsep}{8pt}%
\setlength{\intextsep}{1pt}%
\begin{wrapfigure}{r}{0.20\linewidth}
    \includegraphics[width=\linewidth, trim=660 350 850 270, clip=true]{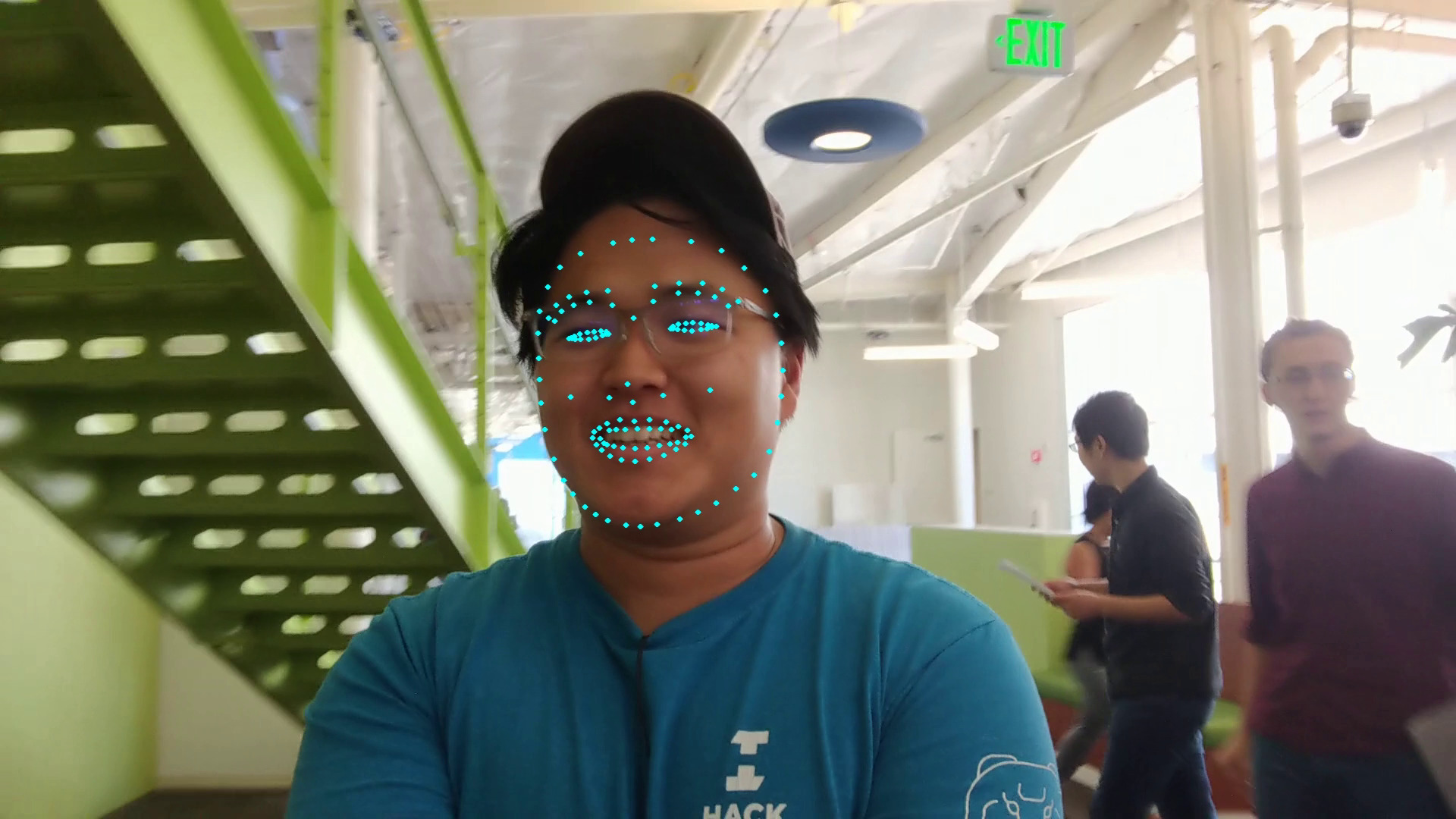}
\end{wrapfigure}
We get the face bounding box from the face detection (several popular ones work equally well in our experiments), and then feed the cropped face into a CNN to obtain the dense landmarks \nothing{\ckliang{Add one image to show the density of our landmarks}}.
The network architecture is a variant of the mobile net \cite{Howard:2017:MEC} which takes $192\times 192\times 3$ input image and returns $133\times 2$ 2D facial landmark coordinates (see inset).
Our GPU implementation can perform at $5.4ms$ \nothing{\fhshi{8.1 - 2.7}} per face on Snapdragon 845.

\textbf{Face of Interest Selection} Due to the time budget, we only select one face to stabilize.
We use a simple yet effective method: for the first frame that contains face(s), we use the largest one returned by the face detection module.
We then keep tracking and stabilizing this face until it is lost.

\subsection{Smooth Head Trajectory Estimation}
In this section, we describe how to obtain a smooth head trajectory from the 2D landmarks, which will be the target head center we want to put the face to.
This smoothing step is critical as some landmarks can have flickering or gross errors.
We model this as an optimization that keeps the head as stable as possible while not causing the virtual frame to move beyond the real frame domain.
The objective function is defined as
\begin{align}
\argmin_\headcenter & w_{1} \ltwo{\headcenter - \headcenter_{\prev}}^2
          + w_{2}\max(\lvert{\headcenter - \landmarkmean}\rvert_{x,y}) / d_{ref})^2, \nonumber \\
&s.t. \lvert{\headcenter - \landmarkmean}\rvert_{x,y} < r,
\end{align}
where $\headcenter$ is the target 2D center on the stabilized virtual frame domain,
$\headcenter_{\prev}$ is the head center of the previous frame,
$\landmarkmean = 1/N \sum_{i}{\landmark_i}$ is the 2D landmark center over the landmarks $\{\landmark_1, ..., \landmark_N\}$,
\nothing{\ckliang{I'm not sure you used L1 or L2 normal for the energy function.}
\fhshi{L2 for both terms, but a max of x and y for the second one.}}
and $d_{ref}$ is a reference deviation that we can tolerant.
If the target head center is not too far away from the real center, the second term would be small.
Thus, $\headcenter$ will tend to be stable as its previous location.
Otherwise, the second term will produce a large penalty and force $\headcenter$ to follow the real landmark center $\landmarkmean$.
$r$ is the cropping ratio at each side of the frame after stabilization.
\nothing{
We set $w_1$ and $w_2$ to 5000 and 1 respectively, and set $d_{ref}$ is to 0.03 \ckliang{r is not set?}.}

\section{Virtual Camera Pose Optimization}
\label{sec:optimization}
With the extracted face information $\{\landmark_i\}$ and the target head center $\headcenter$,
we now describe how to estimate the virtual camera pose so that it fits the face center to the target one, and meanwhile keeps the virtual camera pose changing smoothly across frames.

\subsection{Representation}
Unlike previous images-based methods that use free-form transformations \cite{Grundmann:2011:ADV,Liu:2013:BCP}, we restrict the stabilized camera to have a valid rotation and a shifted projection.
This approach greatly reduces the degree of freedom in optimization and improves the robustness.
We represent the virtual camera pose as a set of 3D rotation and 2D translation:
\begin{equation}
\pose_\virtual = \{\rotation_\virtual, \trans\},
\end{equation}
where $\rotation_\virtual$ is rotation represented by quaternion and $\trans = [t_x, t_y]^T$ is 2D principal offset to the projection center.
Note that the pose, rotation and translation are all functions of time, and we drop the time index for brevity.
\nothing{\ckliang{So, is $P_v$ a function, a set, a matrix, or something else?}
\fhshi{It is a set. Revised.}}

The virtual camera intrinsic matrix is $\intrinsic_\virtual = [\focal_\virtual, 0, 0.5 + t_x;$ $0, \focal_\virtual, 0.5 + t_y; 0, 0, 1]$, where $\focal_\virtual$ is the virtual focal length, which is manually chosen and fixed in our system.

We represent the real camera pose and intrinsic matrix in a similar way:
$\pose_\real = \{\rotation_\real, 0\}$ and $\intrinsic_\real = [\focal_\real, 0, 0.5;$ $0, \focal_\real, 0.5; 0, 0, 1]$.
The real camera does not have principal point shift, and focal length $\focal_\real$ is obtained by calibration.
$\rotation_\real$ is the integration of the angular velocity signal from the gyroscope \cite{Karpenko:2011:DVS}.

Given $\intrinsic_\real$ and $\rotation_\real$ of the current frame, the projection of an image point from the real camera domain to virtual camera domain is decided by a homography transform:
\begin{equation}
\homography(\pose_\virtual, \pose_\real) = \intrinsic_\virtual \Rotation_\virtual (\intrinsic_\real \Rotation_\real)^{-1}.
\end{equation}
where $\Rotation$ is the matrix form of the rotation $\rotation$.
The projection of a 2D facial landmark $\landmark$ from the input image to the stabilized virtual domain is
\begin{equation}
\proj(\landmark, \pose_\virtual, \pose_\real) = \homography\cdot [\landmark_{x}, \landmark_{y}, 1]^T.
\end{equation}
Note that if $\trans = 0$, this transform would only work for objects sufficiently far away from the camera.
The additional principal offset enables us to properly transform close subject like faces.

\subsection{Objective Function}
The goal of stabilization is to find the optimal virtual camera pose $\pose_\virtual$ at each frame.
For real-time viewfinder and streaming applications, we also want to calculate these values without relying on future (non-casual) information.
We cast this process as an optimization problem to minimize the following objective function:
\begin{equation}
\begin{split}
\argmin_{\pose_\virtual} &w_\fit E_{\fit}(\pose_\virtual) + w_\distortion E_{\distortion}(\rotation_\virtual) +
 w_\follow E_{\follow}(\rotation_\virtual) + \\
 & w_\smoothr E_{\smoothr}(\rotation_\virtual) +
  w_\smootht E_{\smootht}(\trans) + w_\protrusion E_{\protrusion}(\pose_\virtual).
\end{split}
\end{equation}
The fixed inputs, such as $\pose_\real$, $\{\landmark_i\}$ and $\headcenter$, are skipped in the argument for brevity.

The \emph{landmark fitting term} $E_{\fit}$ measures the fitting error of the projected landmarks to the target center:
\begin{equation}
E_{\fit}(\pose_\virtual)=\sum_{i} \ltwo{\proj(\landmark_i, \pose_\virtual, \pose_\real) - \headcenter}^2.
\end{equation}

The \emph{distortion term} $E_\distortion$ measures the spherical angle $\solidangle$ between $\rotation_\virtual$ and $\rotation_\real$:
\begin{equation}
E_{\distortion}(\rotation_\virtual) = (\logistic(\solidangle(\rotation_\virtual,\rotation_\real)) \cdot \solidangle(\rotation_\virtual,\rotation_\real))^2.
\end{equation}
A logistic regression function is applied here so that the penalty is close to zero when $\solidangle$ is smaller than a threshold, and increases when $\solidangle$ becomes large.
In other words, this term tolerants the virtual-real camera pose difference within a threshold, and creates large penalty after the difference is further increased.

The \emph{rotation following term} $E_\follow$ measures how the virtual camera follows the real camera.
Unlike the distortion term above, it consistently puts a penalty if the virtual camera rotation is different from the real camera rotation.
The goal is to reduce the change of hitting boundary due to the virtual camera being too stable.
\begin{equation}
E_{\follow}(\rotation_\virtual) = \ltwo{\rotation_\virtual - \rotation_\real}^2.
\end{equation}

The \emph{rotation smoothness term} $E_{\smoothr}$ measures how smooth virtual rotation changes across frames.
It consists of two terms, which controls the C0 and C1 smoothness.
\begin{equation}
\begin{split}
E_{\smoothr}(\rotation_\virtual) & = w_{\smoothr, C0}\ltwo{\rotation_\virtual - \rotation_{\virtual, \prev}}^2 \\
                  & + w_{\smoothr, C1}\ltwo{\rotation_\virtual \rotation_{\virtual, \prev}^{-1} - \rotation_{\virtual, \prev} \rotation_{\virtual, \prevprev}^{-1}}^2,
\end{split}
\end{equation}
where the subscript $_{\prev}$ and $_{\prevprev}$ denote values from the previous and previous-previous frames, respectively.

Similarly, the \emph{translation smoothness term} $E_{\smootht}$ measures how smooth the principal offset changes across frames, which are 
\begin{equation}
E_{\smootht}(\trans) = w_{\smootht, C0} \ltwo{ \trans - \trans_{\prev}}^2
                     + w_{\smootht, C1} \ltwo{2\trans_{\prev} - (\trans + \trans_{\prevprev})}^2.
\end{equation}
\nothing{
\ckliang{No square on the first term?}
\fhshi{Added. Also corrected the smoothness description (L=>C).}
}

Finally, the \emph{protrusion term} $E_{\protrusion}$ measures how the warped frame protrudes the real image boundary:
\begin{equation}
E_{\protrusion}(\pose_\virtual)= \ltwo{\protrude(\pose_\virtual, \pose_\real)/\alpha}^2,
\end{equation}
where $\protrude(\pose_\virtual, \pose_\real)$ is the amount that the warped frame protrudes the real image boundary (we actually make it more sensitive by shrinking the real image boundary to a smaller bounding box), and $\alpha$ is a reference protrusion value we can tolerate (see \Cref{fig:protrusion}).
This concept was introduced for post-processing in \cite{Bell:2014:ANF}, and here we combine it into the joint optimization process.

\subsection{Optimization}
The objective function can be effectively solved by non-linear least square solver such as Ceres \cite{Agarwal:Ceres}.
For the first frame, we initialize the virtual camera pose to identify rotation and zero offset.
For the following frames, we initialize $\rotation_\virtual$ by applying the real camera rotation between current and previous frames to the previous virtual camera rotation, and $\trans_\virtual$ with the previous virtual principal offset:
\begin{equation}
    \pose_\virtual = \{(\rotation_\real \rotation_{\real,\prev}^{-1})\rotation_{\virtual,\prev}, \trans_{\prev}\}.
\end{equation}
The optimization usually converges within 3 iterations and runs at 2.7ms/frame on Snapdragon 845.
\nothing{\ckliang{Say something about the optimization speed.}
\fhshi{Added}}

\subsection{Dynamic Optimization Weight Adjustments}
Up to now, we are able to stabilize the face motion based on 2D landmarks and gyro.
However, there are three practical issues remained.
First, when camera is relatively stable, users are expecting a stable virtual camera.
However, face translations will move the virtual camera around as it follows the face.
Second, the virtual camera will move when user is simply rotating the head.
This is because the 2D landmark center is not pose-invariant.
Finally, the landmarks can be noisy especially when the face rotates away from the camera.
The optimized virtual camera pose will jitter in these cases.

To address these challenges, we dynamically adjust the optimization weights based on gyro and landmarks.
First, we examine the mean magnitude of angular velocity over a period, and adjust $w_\fit$ proportionally.
Next, we decrease the fitting weight and increase the smoothness weights when face pose is large.
Finally, we check the variations of landmark center and scales and decrease the fitting weight if they are large.
As a result, the final virtual camera motion does not move with face when camera is stable or head pose is fast changing, and robust to noisy landmark locations.

\begin{figure}[!t]
    \centering
    \includegraphics[width=0.45\columnwidth]{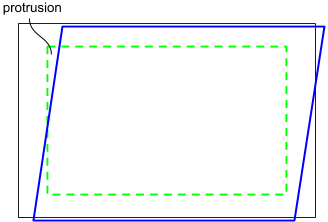}
    \vspace{-0.15in}
    \caption{Protrusion visualization. A protrusion is the amount that the warped video frame (blue) protrudes the the stabilized frame (green) which is cropped from the full frame (black).\label{fig:protrusion}}
\end{figure}

\textbf{Protrusion Handling} The final stabilized frame is a center crop from the warped frame (\Cref{fig:protrusion}), and any protruded area would be undefined.
In rare cases, protrusions can still occur, and we eliminate them by binary searching between $\pose_\virtual$ and $\pose_\real$.
If the binary search failed (no valid solution), we apply the previous warping directly to the current frame.

\begin{figure}[t]
	\centering
	\includegraphics[width=0.95\columnwidth]{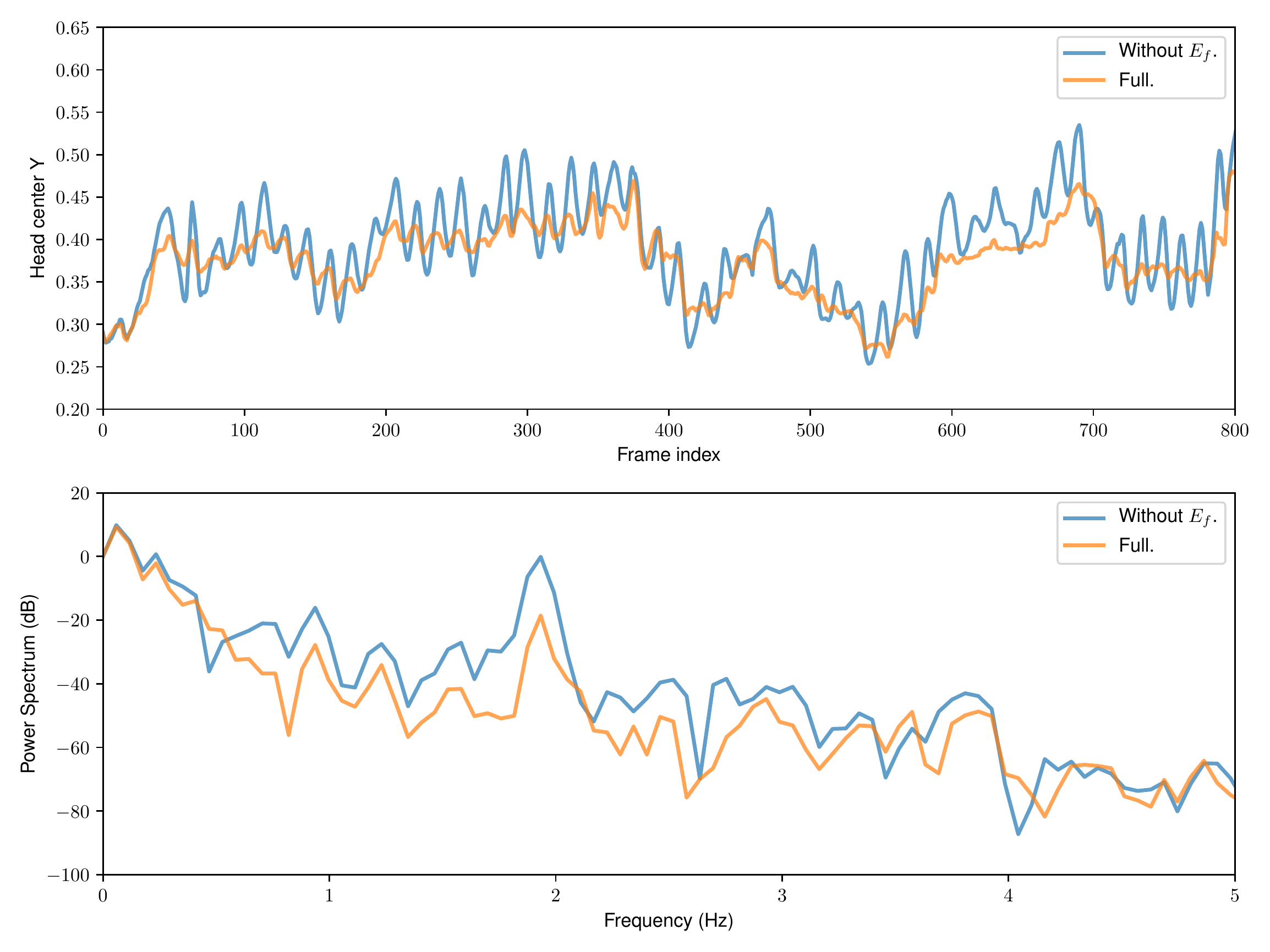}
    \vspace{-0.2in}
	\caption{2D head center trajectory and the power spectrum with/without the fitting term $E_{\fit}$.
        \nothing{\ckliang{Maybe we should also show the power spectrum?}\fhshi{added}}
    } 
	\label{Fig:FittingTermComparison}
\end{figure}

\begin{figure}[t]
	\centering
	\includegraphics[width=0.95\columnwidth]{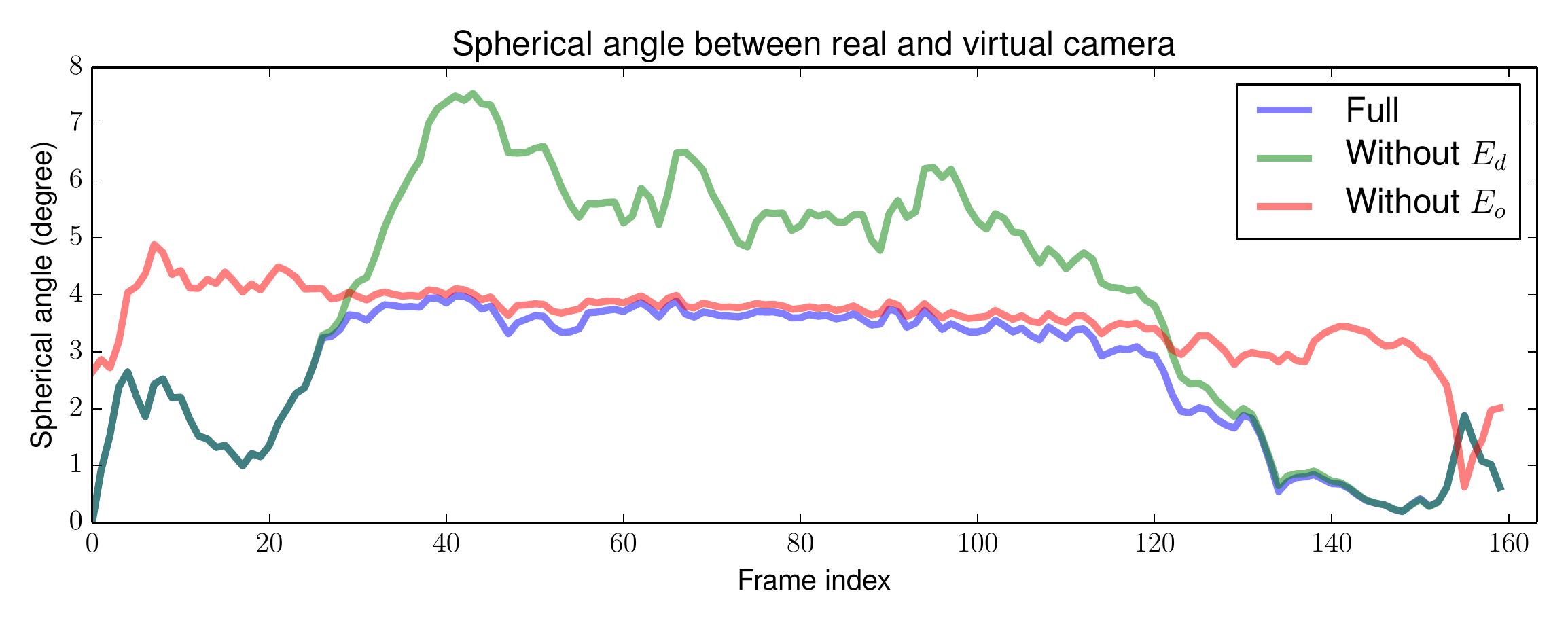}
    \vspace{-0.2in}
	\caption{Virtual camera deviation with/without the rotation following ($E_\follow$) or the distortion ($E_\distortion$) terms.}
	\label{Fig:DistortionFollowingTermComparison}
\end{figure}

\begin{figure}[t]
	\centering
	\includegraphics[width=\columnwidth]{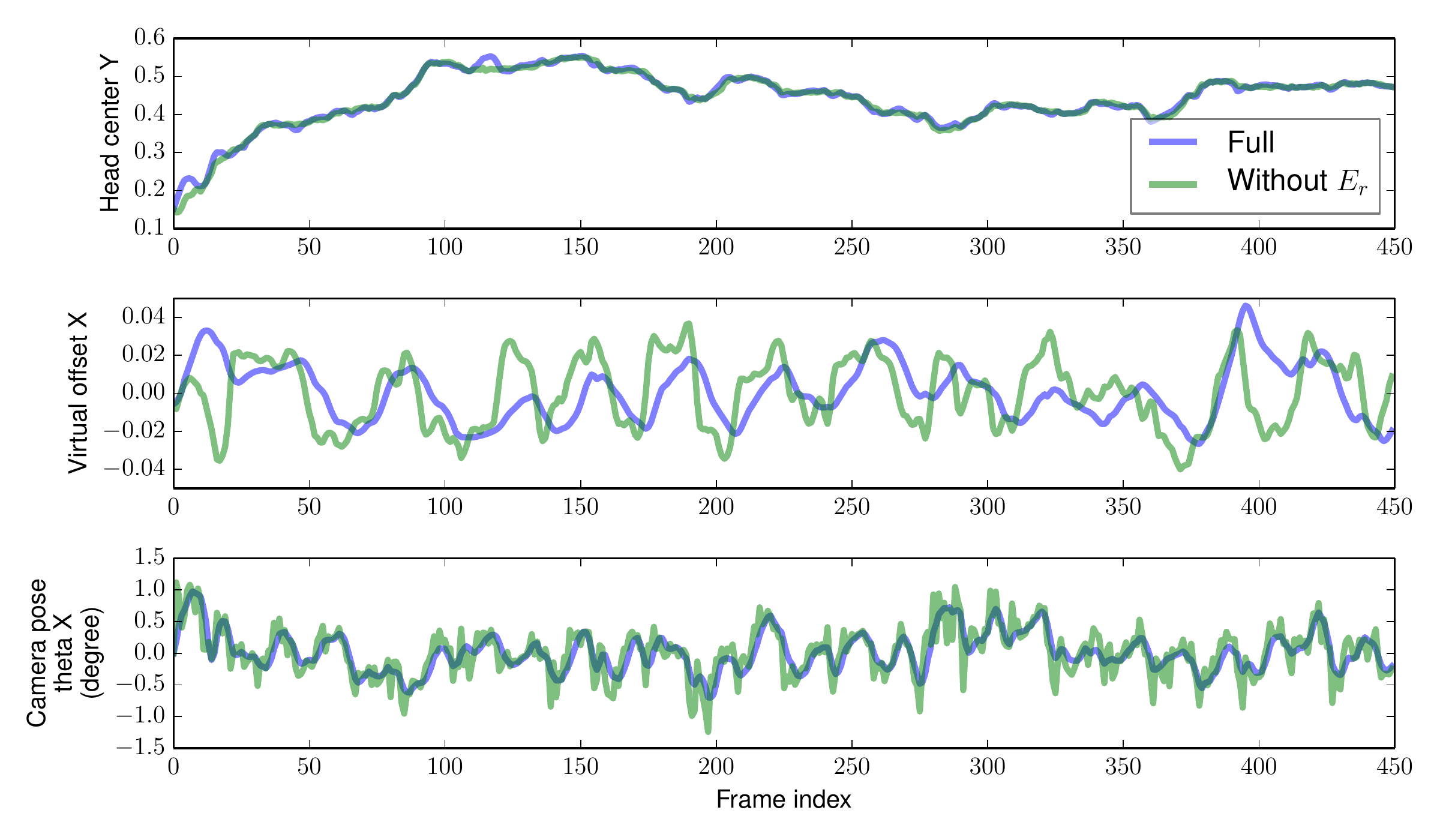}
	\vspace{-0.35in}
    \caption{
        Virtual camera rotation and principal point offset with/without smoothness terms.
        With those terms, the camera (rotation and translation) becomes smoother while the head center (top) remains stable.
        This shows that both the head and background are properly stabilized.}
    \label{Fig:SmoothnessTermComparison}
\end{figure}

\newcommand{\facecrop}[7]{\includegraphics[height=#1\textheight, width=#2\textheight, trim=#3 #4 #5 #6, clip=true]{#7}}

\newcommand{\facecropwithtext}[8]{
	\begin{overpic}[height=#1\textheight, width=#2\textheight, trim=#3 #4 #5 #6, clip=true]{#7}
	\end{overpic}
}

\newcommand{\facetriplet}[9]{
    \facecropwithtext{0.06}{0.055}{#1}{#2}{#3}{#4}{curr_vs_eis3/#7}{#9} &\hspace{-4mm}
    \facecrop{0.06}{0.055}{#5}{#2}{#6}{#4}{curr_vs_eis3/#8} &\hspace{-4mm}
    \facecrop{0.06}{0.055}{#1}{#2}{#3}{#4}{curr_vs_eis3/#8}
}

\begin{figure}[t]
	\centering
    \hspace{-3.1mm}
	\begin{tabular}[t]{cccccc}
        \facetriplet{310}{150}{1250}{5}{1270}{290}{indoor_walk_input.jpg}{indoor_walk.jpg}{v0} &\hspace{-4mm}
        \facetriplet{310}{115}{1250}{40}{1295}{285}{outdoor_walk_input.jpg}{outdoor_walk.jpg}{v4}\\

        \facetriplet{285}{175}{1360}{75}{1260}{385}{outdoor_panning2_input.jpg}{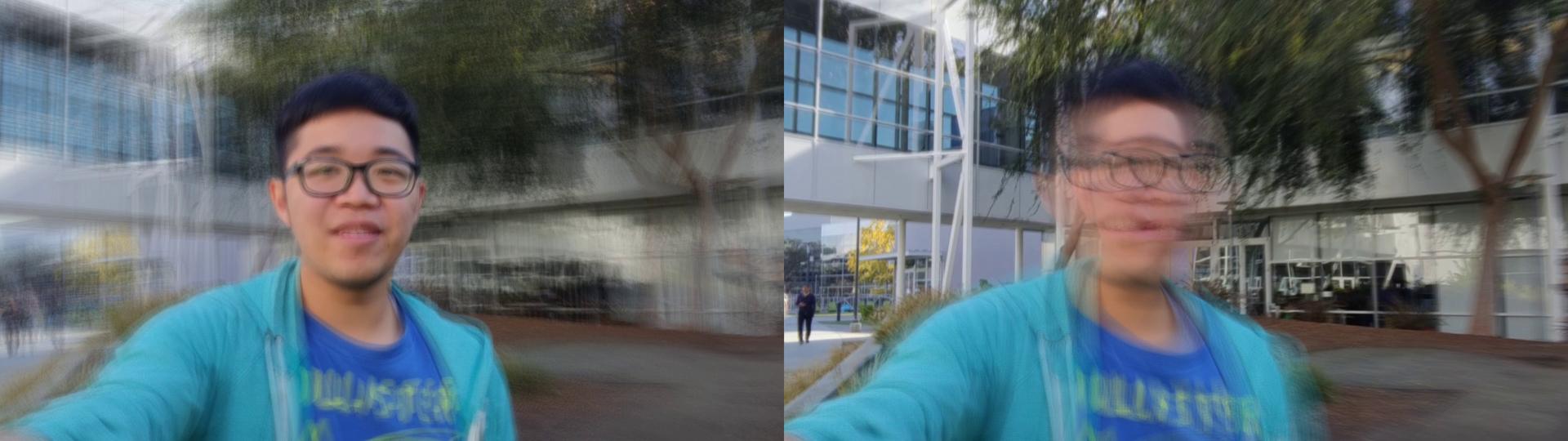}{v6} &\hspace{-4mm}
        \facetriplet{275}{150}{1350}{50}{1235}{390}{outdoor_panning1_input.jpg}{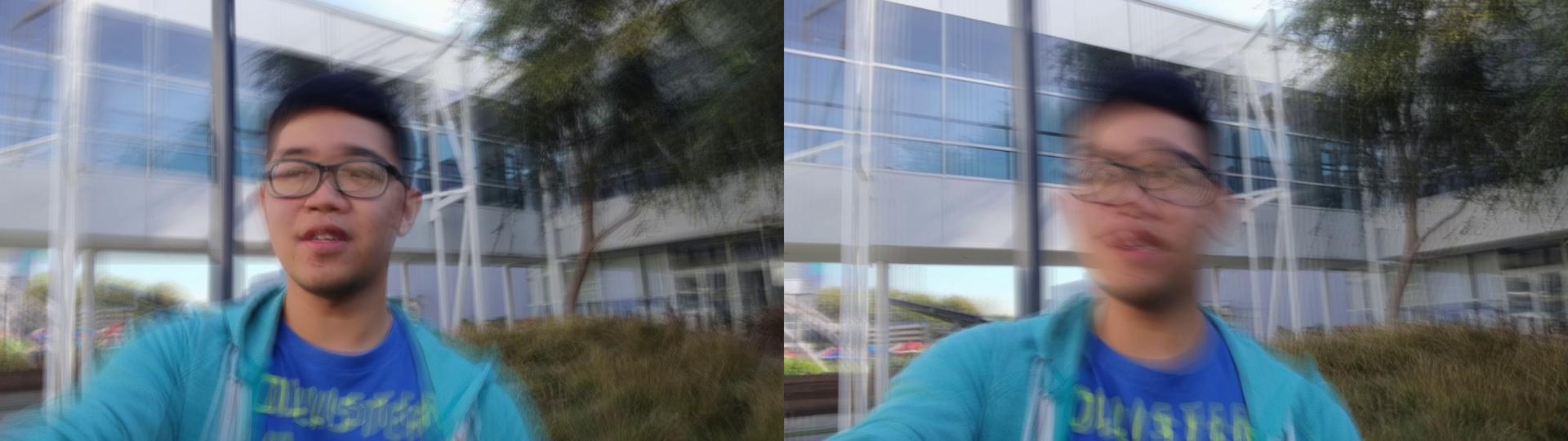}{v7}\\

        \facetriplet{310}{145}{1270}{40}{1270}{310}{outdoor_walk1_input.jpg}{outdoor_walk1.jpg}{v3} &\hspace{-4mm}
        \facetriplet{310}{162}{1270}{0}{1270}{310}{outdoor_walk2_input.jpg}{outdoor_walk2.jpg}{v3}\\

        \facetriplet{315}{135}{1267}{55}{1275}{307}{outdoor_panning_input.jpg}{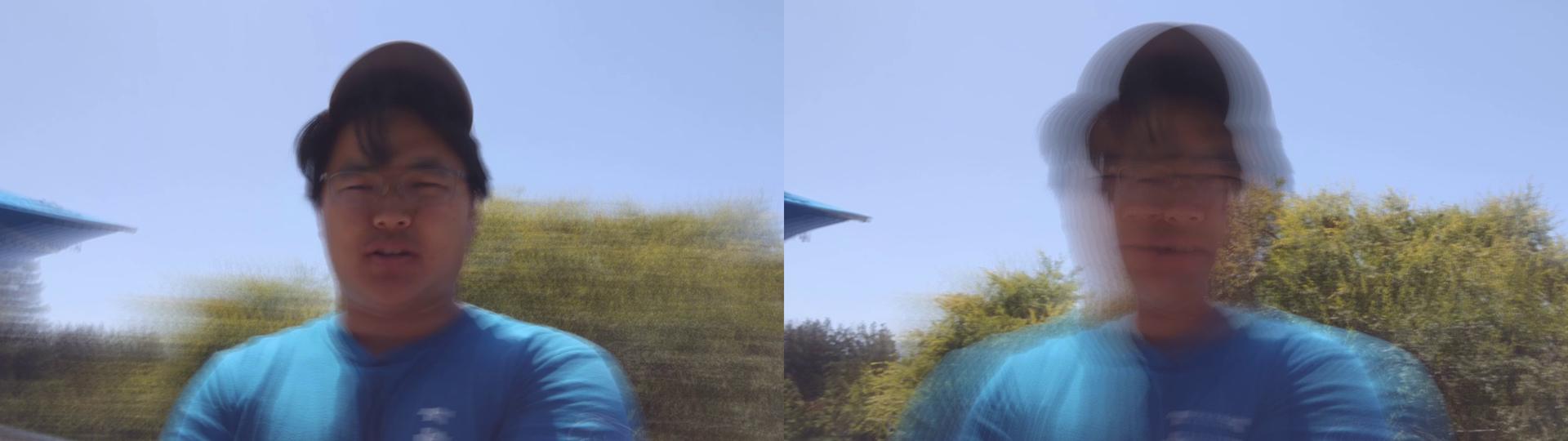}{v5} &\hspace{-4mm}
        \facetriplet{135}{105}{1395}{0}{1095}{435}{outdoor_occlusion_input.jpg}{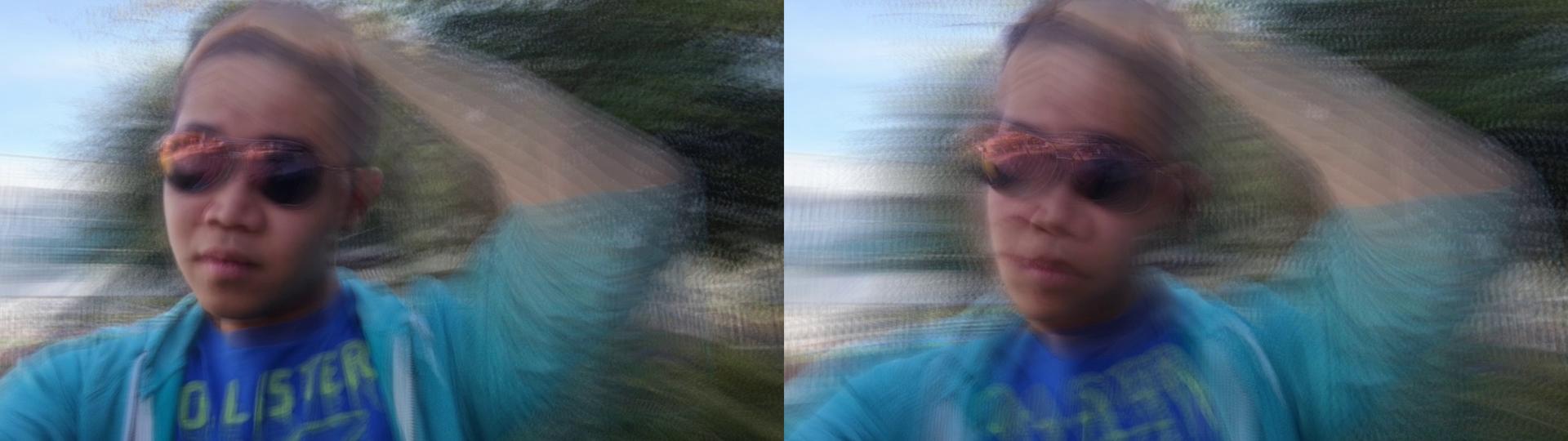}{v1}\\
        \vspace{-0.05in}
		(a)&(b)&(c)&(a)&(b)&(c)
	\end{tabular}
    \vspace{-0.1in}
	\caption{
        Head stabilitiy visualization. Each frame is the average of consecutive 15 frames.        
        For each triplet: (a) input video with video id in the accompanying video \cite{Shi:2019:SUP}, (b) results by the fused video stabilization \cite{Liang:2017:FVS}, and (c) our results.}
	\label{Fig:SteadifaceVSEIS3}
\end{figure}
\renewcommand{\facecrop}[7]{\includegraphics[height=#1\textheight, width=#2\textheight, trim=#3 #4 #5 #6, clip=true]{#7}}

\renewcommand{\facetriplet}[9]{
    \facecropwithtext{0.06}{0.055}{#1}{#2}{#3}{#4}{curr_vs_eccv/#7}{#9} &\hspace{-4mm}
    \facecrop{0.06}{0.055}{#1}{#5}{#3}{#6}{curr_vs_eccv/#8}
}

\begin{figure}[t]
	\centering
    \hspace{-3.3mm}
	\begin{tabular}[t]{cccccc}
        \facetriplet{620}{300}{600}{1090}{1380}{10}{outdoor_walk2.jpg}{outdoor_walk2.jpg}{v4} &\hspace{-5mm}
        \facetriplet{700}{300}{540}{1130}{1380}{40}{outdoor_walk1.jpg}{outdoor_walk1.jpg}{v3} &\hspace{-5mm}
        \facetriplet{450}{180}{640}{1090}{1260}{10}{outdoor_walk.jpg}{outdoor_walk.jpg}{v10}\\

        \facetriplet{800}{500}{400}{1090}{1480}{10}{outdoor_still.jpg}{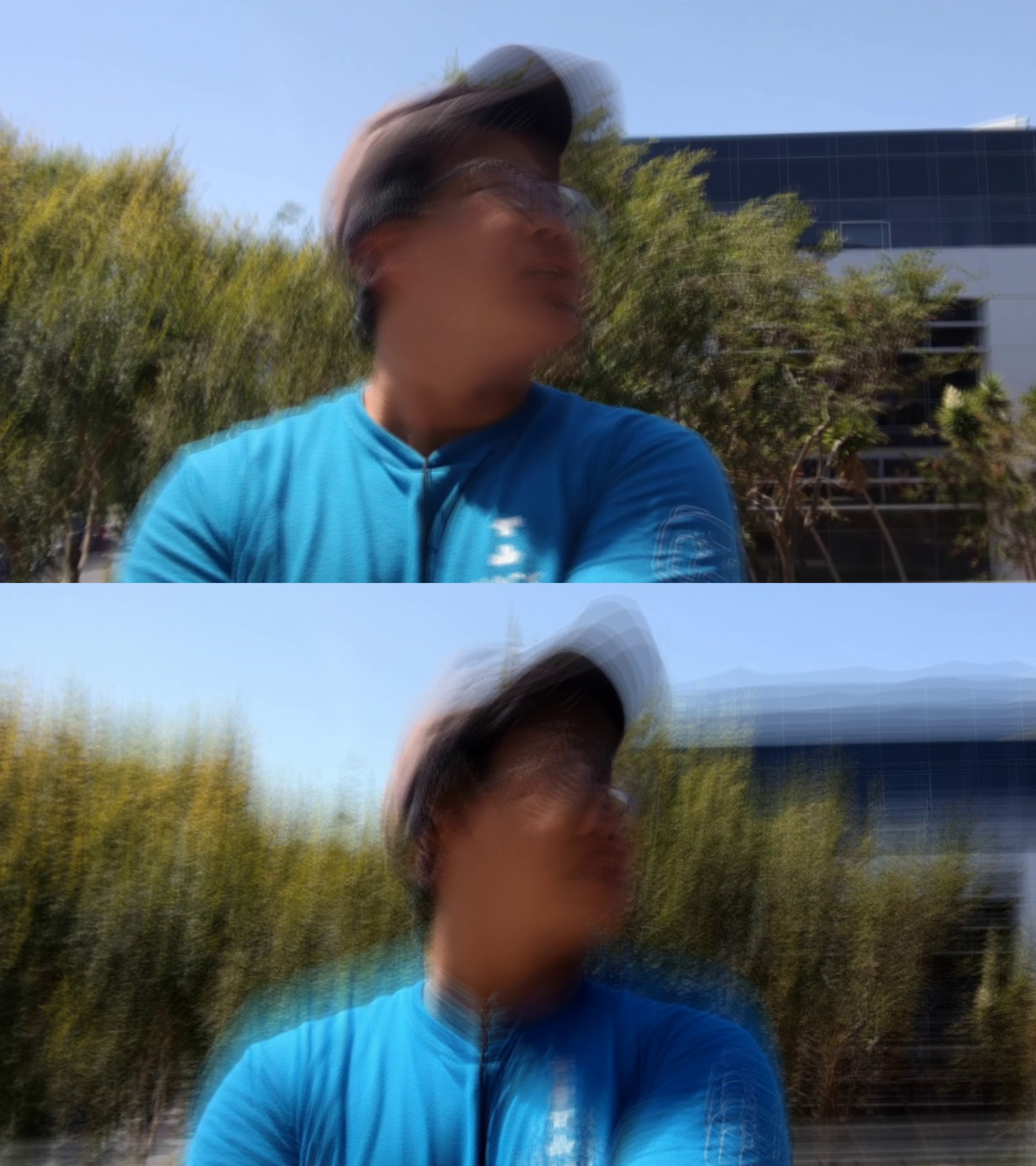}{v8} &\hspace{-5mm}
        \facetriplet{500}{300}{800}{1100}{1380}{20}{indoor_walk.jpg}{indoor_walk.jpg}{v0} &\hspace{-5mm}
        \facetriplet{500}{300}{800}{1100}{1380}{20}{indoor_panning.jpg}{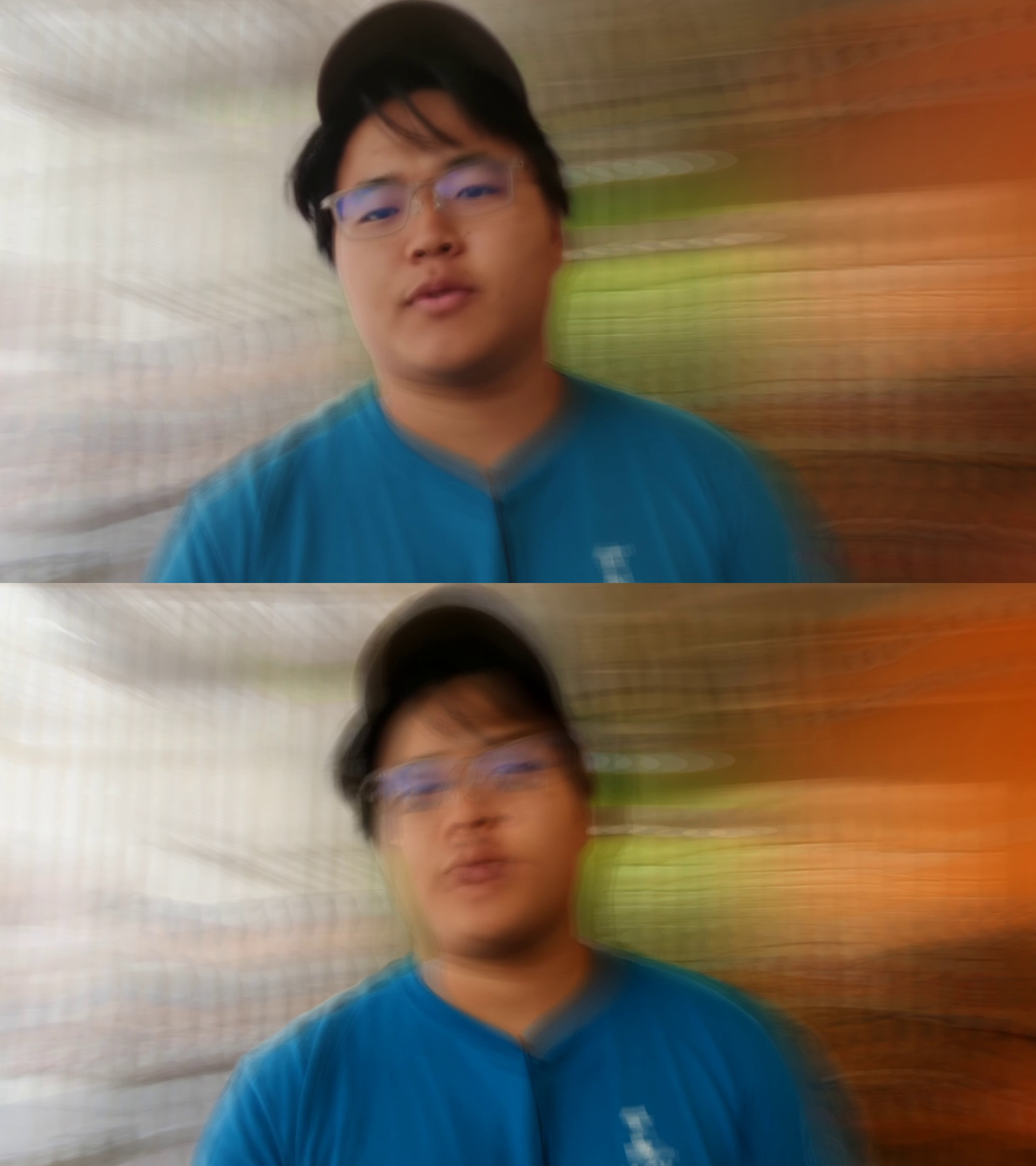}{v9}\\
        \vspace{-0.05in}
		(a)&(b)&(a)&(b)&(a)&(b)
	\end{tabular}
    \vspace{-0.1in}
	\caption{
        Comparison against video stabilization method in \cite{Yu:2018:SVS}. Each frame is the average of consecutive 15 frames.
        For each pair: (a) results by Yu and Ramamoorthi \cite{Yu:2018:SVS} with video id in the accompanying video \cite{Shi:2019:SUP}, and (b) our results.}
	\label{Fig:SteadifaceVSECCV}
\end{figure}

\section{Results}
\label{sec:results}

In this section, we first demonstrate the effectiveness of our method on videos with a variety of head and hand motions.
We show our method stabilizes both the head motion and the background, and is robust to large head pose variations (e.g. $v0$, $v7$ in the accompanying video \cite{Shi:2019:SUP}), illumination changes (e.g. $v0$, $v5$) and occlusions (e.g. $v1$, $v2$).
We then validate the method by evaluating the importance of each term.
Next, we compare our method with the state-of-the-art gyro-based stabilization on mobile phone \cite{Liang:2017:FVS}.
Finally, we compare our method with the state-of-the-art selfie video stabilization method \cite{Yu:2018:SVS}.
Our results show comparable or better face stability, and do not suffer from artifacts caused by unreliable optical flow/landmarks.

We tested our method on 43 \nothing{\fhshi{number here}}videos with a combination of different head motions, expressions (\eg talking to camera, looking around) and hand motions (\eg tripod, walking, panning).
All results are generated with the identical parameter set, and they are best seen in the accompanying video at \cite{Shi:2019:SUP}.

\subsection{Importance of Each Term}

We show the importance of each term by disabling them during the optimization.
\Cref{Fig:FittingTermComparison} shows the head trajectory and power spectrum with/without the fitting term $E_{\fit}$.
As we can see, the head remains unstable when stabilizing using only gyro, and the fitting term effectively reduces the head motion and produces a lower power density over frequencies.
Note that the stabilized trajectory is not perfectly smooth as it is a trade-off between the fitting term and other terms.
\nothing{\fhshi{TODO: refine the fonts in figure, and may replace \Cref{Fig:FittingTermComparison} with a stat table.}}

\Cref{Fig:DistortionFollowingTermComparison} shows the deviation between the virtual camera pose and the real camera pose during fast panning.
Clearly, the rotation following term $E_\follow$ makes the virtual pose well defined when multiple solutions exist and the solution that is close to the real pose is selected.
Meanwhile, the distortion term $E_\distortion$ further refines the solution space by adaptively imposing penaulty when the real-virtual rotation deviation tends to be large.

Finally, the \Cref{Fig:SmoothnessTermComparison} shows the rotation and principal offset curves with/without the smoothness terms which demonstrates their necessity for balancing the head and background stability.
\nothing{\ckliang{This one is not very strong to me. In the end what matters is the head center smoothness. Smoother $\pose_\virtual$ is always better?}}

\subsection{Comparison to State-of-the-art Gyro-based Method}
We now compare our method with the fused video stabilization (FVS) \cite{Liang:2017:FVS} on Google Pixel 2/3.
It is rated as one of best video stabilization solutions on mobile phones by many reviews.
We show the stabilization effect by averaging the consecutive 15 frames (0.5s) in \Cref{Fig:SteadifaceVSEIS3}.

As we can see, the face and background are blurry in the inputs due to shaky motions.
FVS stabilizes the background, but exaggerates the foreground head motion.
In contrast, our \emph{Steadiface} outputs stable head motion across frames while maintaining a good trade-off for background stability.
Note that our method uses a smaller cropping ratio ($15\%$) than FVS does ($20\%$).
This makes stabilization more challenging, but we can preserve more field-of-view for users.

\subsection{Comparison to State-of-the-art Image-based Method}
Finally, we compare \emph{Steadiface} with the selfie video stabilization method \cite{Yu:2018:SVS} (\Cref{Fig:SteadifaceVSECCV}).
Note that their solution cannot reach real-time performance even on a desktop.
The face stability of \emph{Steadiface} is comparable or slightly better than them.
Meanwhile, \emph{Steadiface} does not suffer from quick jittering when landmark detection/optical flow fails (e.g. the shaky background at bottom left of \Cref{Fig:SteadifaceVSECCV}).
One drawback is the cropping ratio.
Their method dynamically adjusts the crop and can preserve wider field-of-view sometimes.

In sum, we present \emph{Steadiface}, a new real-time face-centric video stabilization method that simultaneously removes hand shake and keeps subject's head stable.
It can work with different types of camera motions, and is robust to large head pose, occlusion and facial appearance variations.
It is also highly efficient and runs at 8.1 ms/frame with a single core on Google Pixel 3.

\nothing{\ckliang{Discussion on implementation platform and runtime somewhere.}}

{ 
    \footnotesize \vfill
    \bibliographystyle{IEEEbib} 
    \bibliography{main}

\begin{thebibliography}{10}

\bibitem{Shi:2019:SUP}
``Supplemental videos,''
  \url{https://drive.google.com/open?id=1nSaPwAsrtkY3bhd1S35GoJM4Mh5uK1IQ}.

\bibitem{Morimoto:1998:EIS}
C.~Morimoto and R.~Chellappa,
\newblock ``Evaluation of image stabilization algorithms,''
\newblock in {\em ICASSP}, 1998, vol.~5, pp. 2789--2792.

\bibitem{Karpenko:2011:DVS}
A.~Karpenko, D.~Jacobs, J.~Baek, and M.~Levoy,
\newblock ``Digital video stabilization and rolling shutter correction using
  gyroscopes,''
\newblock {\em CSTR}, 2011.

\bibitem{Grundmann:2011:ADV}
M.~Grundmann, V.~Kwatra, and I.~Essa,
\newblock ``Auto-directed video stabilization with robust {L1} optimal camera
  paths,''
\newblock in {\em CVPR}, 2011.

\bibitem{Liang:2008:ACR}
C.-K. Liang, L.-W. Chang, and H.~H. Chen,
\newblock ``Analysis and compensation of rolling shutter effect,''
\newblock {\em IEEE TIP}, vol. 17, no. 8, 2008.

\bibitem{Bell:2014:ANF}
S.~Bell, A.~Troccoli, and K.~Pulli,
\newblock ``A non-linear filter for gyroscope-based video stabilization,''
\newblock in {\em ECCV}, 2014.

\bibitem{Liang:2017:FVS}
``Fused video stabilization on the {Pixel} 2 and {Pixel} 2 {XL},''
  \url{https://ai.googleblog.com/2017/11/fused-video-stabilization-on-pixel-2.html},
\newblock Accessed: 2018-12-23.

\bibitem{Liu:2009:CPW}
F.~Liu, M.~Gleicher, H.~Jin, and A.~Agarwala,
\newblock ``Content-preserving warps for 3{D} video stabilization,''
\newblock {\em ACM TOG}, vol. 28, no. 3, 2009.

\bibitem{Goldstein:2012:VSE}
A.~Goldstein and R.~Fattal,
\newblock ``Video stabilization using epipolar geometry,''
\newblock {\em ACM TOG}, vol. 31, no. 5, 2012.

\bibitem{Liu:2011:SVS}
F.~Liu, M.~Gleicher, J.~Wang, H.~Jin, and A.~Agarwala,
\newblock ``Subspace video stabilization,''
\newblock {\em ACM TOG}, vol. 30, no. 1, 2011.

\bibitem{Liu:2013:BCP}
S.~Liu, L.~Yuan, P.~Tan, and J.~Sun,
\newblock ``Bundled camera paths for video stabilization,''
\newblock {\em ACM TOG}, vol. 32, no. 4, 2013.

\bibitem{Yu:2018:SVS}
J.~Yu and R.~Ramamoorthi,
\newblock ``Selfie video stabilization,''
\newblock in {\em ECCV}, 2018.

\bibitem{Blanz:1999:Morphable}
V.~Blanz and T.~Vetter,
\newblock ``A morphable model for the synthesis of 3{D} faces,''
\newblock in {\em SIGGRAPH}, 1999, pp. 187--194.

\bibitem{Shi:2014:Automatic}
F.~Shi, H.-T. Wu, X.~Tong, and J.~Chai,
\newblock ``Automatic acquisition of high-fidelity facial performances using
  monocular videos,''
\newblock {\em ACM TOG}, vol. 33, no. 6, pp. 222, 2014.

\bibitem{Thies:2016:face2face}
J.~Thies, M.~Zollhofer, M.~Stamminger, C.~Theobalt, and M.~Nie{\ss}ner,
\newblock ``Face2face: Real-time face capture and reenactment of rgb videos,''
\newblock in {\em CVPR}, 2016, pp. 2387--2395.

\bibitem{Garrido:2016:3DFaceRig}
P.~Garrido, M.~Zollh{\"o}fer, D.~Casas, L.~Valgaerts, K.~Varanasi,
  P.~P{\'e}rez, and C.~Theobalt,
\newblock ``Reconstruction of personalized 3{D} face rigs from monocular
  video,''
\newblock {\em ACM TOG}, vol. 35, no. 3, pp. 28, 2016.

\bibitem{Cao:2014:DDE}
C.~Cao, Q.~Hou, and K.~Zhou,
\newblock ``Displaced dynamic expression regression for real-time facial
  tracking and animation,''
\newblock {\em ACM TOG}, vol. 33, no. 4, pp. 43, 2014.

\bibitem{Jourabloo:2016:CNN}
A.~Jourabloo and X.~Liu,
\newblock ``Large-pose face alignment via {CNN}-based dense 3{D} model
  fitting,''
\newblock in {\em CVPR}, 2016, pp. 4188--4196.

\bibitem{Tewari:2017:Mofa}
A.~Tewari, M.~Zollh{\"o}fer, H.~Kim, P.~Garrido, F.~Bernard, P.~P{\'e}rez, and
  C.~Theobalt,
\newblock ``Mofa: Model-based deep convolutional face autoencoder for
  unsupervised monocular reconstruction,''
\newblock in {\em ICCV}, 2017, vol.~2, p.~5.

\bibitem{Howard:2017:MEC}
A.~G. Howard, M.~Zhu, B.~Chen, D.~Kalenichenko, W.~Wang, T.~Weyand,
  M.~Andreetto, and H.~Adam,
\newblock ``Mobilenets: Efficient convolutional neural networks for mobile
  vision applications,''
\newblock {\em arXiv preprint arXiv:1704.04861}, 2017.

\bibitem{Agarwal:Ceres}
S.~Agarwal, K.~Mierle, and Others,
\newblock ``Ceres solver,'' \url{http://ceres-solver.org},
\newblock Accessed: 2018-12-23.

\end{thebibliography}
}

\end{document}